\documentclass[a4paper, 10 pt, conference]{cras} 
\usepackage[utf8]{inputenc}
\IEEEoverridecommandlockouts                       
\usepackage{mathtools}
\usepackage{geometry}
\usepackage{newtxmath,newtxtext}
\usepackage{authblk}
\usepackage{pgfplots}
\usepackage{todonotes}
\usepackage{subfig}
\usepackage{graphicx}
\usepackage{siunitx}

\pgfplotsset{compat=newest} 
\pgfplotsset{plot coordinates/math parser=false} 
 
\title{\Large \bf
Design of a Polymer-based Steerable Cannula for Neurosurgical Applications} 

\author{\large Nidhi Malhotra*}
\author{\large Amber K. Rothe} 
\author{\large Revanth Konda} 
\author{\large Jaydev P. Desai} 
\vspace{10pt}
\affil{\small\textit{Medical Robotics and Automation (RoboMed) Laboratory, Georgia Institute of Technology}}
\affil{\small\textit{*Corresponding Author, Email: nmalhotra37@gatech.edu}}

\begin{document}

\maketitle
\thispagestyle{empty}
\pagestyle{empty}

\section*{INTRODUCTION}
Robotically steerable compliant surgical tools offer several advantages over rigid tools, including enhanced dexterity, reduced tissue damage, and the ability to generate non-linear trajectories in minimally invasive neurosurgical procedures \cite{Liu2024}. Many existing robotic neurosurgical tools are designed using stainless steel or nitinol materials \cite{Qi2025, Chitalia2020}. Using polymer-based materials instead can offer advantages such as reduced interference in magnetic resonance imaging \cite{Liu2024}, enhanced safety for guiding electrically powered instruments \cite{Gao2019,Leber2023}, and reduced tissue damage due to inherent compliance \cite{Gao2019,Leber2023}. Several polymer materials have been used in robotic surgical applications, such as polyimide \cite{Gao2019}, polycarbonate \cite{Leber2023}, and elastic resin \cite{Barnes2023}. Various fabrication strategies have also been proposed, including standard microfabrication techniques \cite{Gao2019}, thermal drawing \cite{Leber2023}, and 3-D printing \cite{Barnes2023}. In our previous work, a tendon-driven, notched-tube was designed for several neurosurgical robotic tools \cite{Qi2025, Chitalia2020}, utilizing laser micromachining to reduce the stiffness of the tube in certain directions. This fabrication method is desirable because it has a single-step process, has high precision, and does not require a cleanroom or harsh chemicals. \textcolor{black}{Past studies have explored laser-micromachining of polymer material for surgical applications such as stent fabrication \cite{grabow}}. In this work, we explore extending the use of the laser-micromachining approach to the fabrication of polyimide (PI) robotically steerable cannulas for neurosurgical applications. Utilizing the method presented in this work, we fabricated joints as small as \SI{1.5}{\milli\meter} outer diameter (OD). Multiple joints were fabricated using PI tubes of different ODs\textcolor{black}{, and the loading behavior of the fabricated joints was experimentally characterized}.

\section*{MATERIALS AND METHODS}
\begin{figure*}[t]
    \centering
    \includegraphics[width=0.83\textwidth]{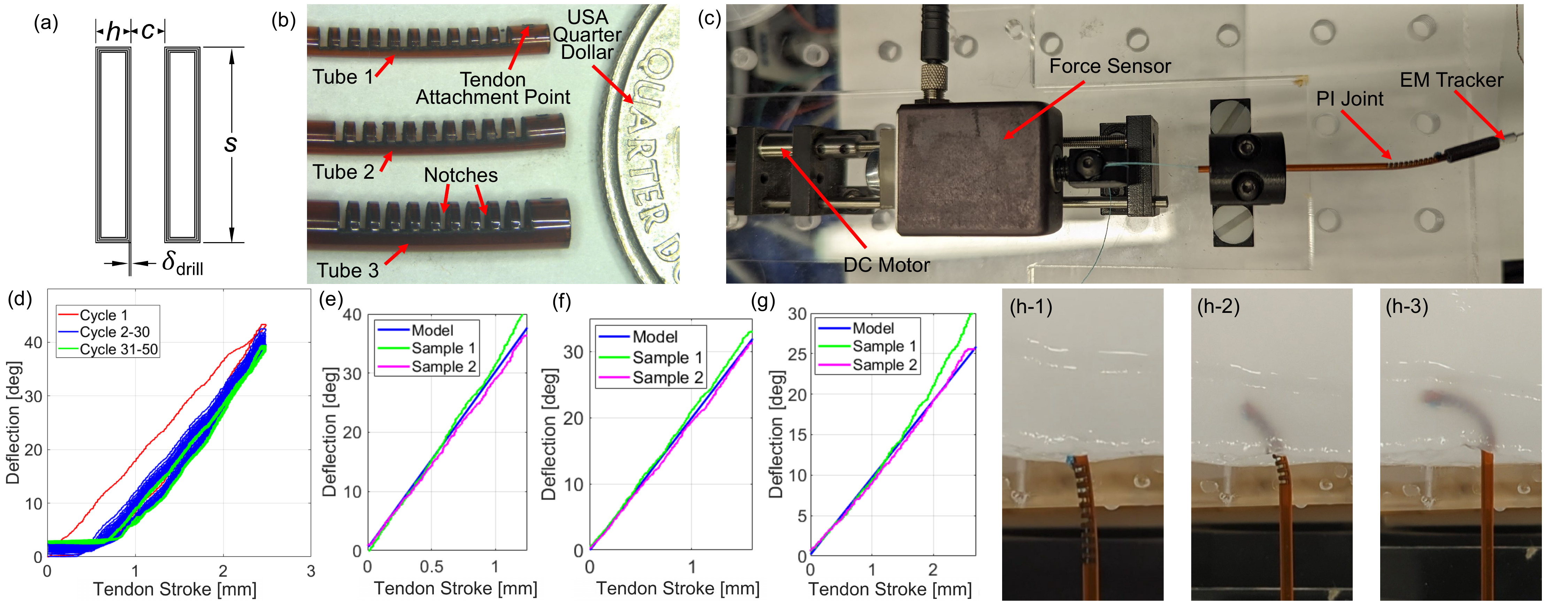}
    \caption{\small (a) Two sample notches with dimensions from Table \ref{table:tube_parameters}. (b) Three micromachined PI tubes and a USA quarter dollar for scale. (c) Experimental setup for joint model validation. (d) Joint deflection vs. tendon stroke for cycle 1 is shown in red, cycle 2-30 is shown in blue, and cycle 31-50 is shown in green. Joint model and experimental data of two samples for (e) Tube 1, (f) Tube 2, and (g) Tube 3. (h-1)-(h-3) Steps of insertion and steering of the PI joint in the hydrogel phantom.\normalsize}
    \label{fig:results}
    \vspace{-6mm}
\end{figure*}

Patterns of unidirectional asymmetric notches (UAN) for machining three differently sized PI tubes were designed in Autocad\textsuperscript{\textregistered} 2025 (Autodesk\textsuperscript{\textregistered}, San Francisco, CA, USA). The dimensions of the tubes and the notch pattern are given in Table\,\ref{table:tube_parameters} and labeled in Fig.\,\ref{fig:results}\,(a). The designs were laser micromachined using a femtosecond laser (WS-Flex Ultra-Short Pulse Laser Workstation, Optec, Frameries, Belgium). Two samples of each of the three tubes were machined. The laser parameters used to cut the notches are provided in Table\,\ref{table:laer_parameters}, and the machined tubes are shown in Fig.\,\ref{fig:results}\,(b). Multiple passes of the laser were used to cut the material without causing significant thermal degradation. Each notch was first cut four times, with a small offset of \(\delta_{\mathrm{drill}}\) between cuts. This procedure was repeated twice: once in focus, and once defocused \(\delta_{\mathrm{defocus}}\) inside the material. Finally, both previous steps were repeated 4, 6, and 8 times for Tube 1, Tube 2, and Tube 3, respectively, depending on the thickness of the material. 
In addition, a \SI{0.3}{}\(\times\)\SI{0.3}{\milli\meter} square hole was cut using the same procedure near each tube's tip to allow attachment of the tendon. An ultra-high molecular weight polyethylene (UHMWPE) tendon with a radius ($r_t$) of \SI{0.04}{\milli\meter} was used to actuate the joints using a DC Motor (Maxon\textsuperscript{\textregistered} Group, Sachseln, Switzerland)  with a 64:1 gear ratio connected to a lead screw as shown in Fig.\,\ref{fig:results}\,(c). 

\begin{table}[t]
\centering
\small
\caption{Tube Geometric Parameters}
\label{table:tube_parameters}
\vspace{-0.25cm}
\begin{tabular}{|c||c|c|c|}
\hline
\textbf{Parameters} & \textbf{Tube 1} & \textbf{Tube 2} & \textbf{Tube 3} \\
\hline

\textbf{Outer Radius (\(r_o\)) [\SI{}{\milli\meter}]} & 0.75 & 0.85 & 1.25\\
\hline
\textbf{Inner Radius (\(r_i\)) [\SI{}{\milli\meter}]} & 0.45 & 0.55 & 0.75\\
\hline
\textbf{Notch Arc (\(s\)) [\SI{}{\milli\meter}]} & 2.83 & 3.20 & 4.71\\
\hline
\textbf{Notch Width (\(h\)) [\SI{}{\milli\meter}]} & 0.50 & 0.50 & 0.50\\
\hline
\textbf{Notch Spacing (\(c\)) [\SI{}{\milli\meter}]} & 0.50 & 0.50 & 0.50\\
\hline
\textbf{Number of Notches (\(n\)) } & 10 & 10 & 10 \\
\hline
\end{tabular}
\vspace{-0.25cm}
\end{table}
\normalsize


\begin{table}[t]
\centering
\small
\caption{Laser Parameters}
\label{table:laer_parameters}
\vspace{-0.25cm}
\begin{tabular}{|c||c|}
\hline
\textbf{Power (\(P\)) [\SI{}{\watt}]} & 2.0\\
\hline
\textbf{Pulse Frequency (\(f\)) [\SI{}{\kilo\hertz}]} & 60  \\
\hline
\textbf{Scan Speed (\(v\)) [\SI{}{\milli\meter\per\second}]} & 25 \\
\hline
\textbf{Wavelength (\(\lambda)\) [\SI{}{\nano\meter}]} & 1030 \\
\hline
\textbf{Drill Offset (\(\delta_{\mathrm{drill}})\) [\SI{}{\milli\meter}]} & 0.005 \\
\hline
\textbf{Defocus Offset (\(\delta_{\mathrm{defocus}})\) [\SI{}{\milli\meter}]} & 0.1 \\
\hline
\end{tabular}
\vspace{-0.8cm}
\end{table}
\normalsize
%
To obtain a relationship between the tendon stroke ($L_{t}$) and the deflection ($\theta$) of the joint, similar to \cite{Qi2025, Chitalia2020}, we derived a model incorporating the kinematic ($L_{kin}$) and tendon elongation ($L_{el}$) effects, such that $L_t=L_{kin}+L_{el}$. The  kinematic term for a UAN joint is given by:

\vspace{-0.5cm}
\small\begin{equation}
    L_{kin}=(\bar{y}+r_{i}-r_t)\theta\quad\text{and}\quad\bar{y}=\frac{4\sin(\frac{\phi}{2})(r_{o}^3-r_{i}^{3})}{3\phi(r_{o}^{2}-r_{i}^{2})}
\end{equation}\normalsize
\vspace{-0.25cm}

\noindent where $\phi$ denotes the angle subtended by the material left after laser micromachining, given by $\phi=\frac{2\pi r_o - s}{r_o}$. The tendon elongation is given by $L_{el}= \frac{FL_{0}}{E_{t}\pi r_{t}^2}$, where, $F$ is the tendon tension, $L_0$ is the length of the tendon when unactuated, and $E_t$ is the Young's modulus of the tendon.

\section*{RESULTS}
Firstly, data for fifty cycles of actuation and relaxation of the joints was collected. The maximum tendon stroke for the Tube 1, Tube 2, and Tube 3 joints was set at \SI{2}{\milli\meter}, \SI{2.5}{\milli\meter}, and \SI{5}{\milli\meter}, respectively. A force sensor (MDB-2.5, Transducer Techniques\textsuperscript{\textregistered}, Temecula, CA, USA) was utilized to measure the tendon tension ($F$), required for joint deflection (Fig.\,\ref{fig:results}\,(c)). An electromagnetic (EM) tracking system (Northern Digital Inc., Aurora, Waterloo, ON, Canada) was used to obtain the joint deflection. As an example, the joint deflection vs. tendon stroke behavior of Sample 1 for Tube 2 is presented in Fig.\,\ref{fig:results}\,(d), where the last 20 cycles are highlighted in green. The plot shows that the first input-output cycle (shown in red) is significantly different compared to the subsequent cycles, a common phenomenon reported for many soft materials \cite{konda2022}. Furthermore, we observe slight variations during the subsequent cycles (hypothesized due to creep in the material); however, negligible variation in the joint deflection was observed in the last 20 cycles. For each tube, the experimental and modeled deflection of the two samples for the last loading cycle are shown in Fig.\,\ref{fig:results}\,(e)-(g). The deadband from the tendon slack (caused due to the variations in the cycles) has been removed in these plots. The undeformed length of the tendon ($L_0$) was measured to be 70\,mm, and the Young's modulus ($E_t$) of the tendon was estimated to be \SI{28}{\giga\pascal}, since this resulted in minimum error between the modeled and experimental data. The model estimated the joint angles with root-mean-square-error (RMSE) values given in Table\,III. For both the samples of Tube 1 and Tube 2 and sample 2 of Tube 3, the RMSE-values between the joint deflection estimated from the model and experimental data are less than 1$^\circ$. However, for sample 1 of Tube 2, we observe a slight deviation from the model at higher tendon stroke values. Finally, the steering of Tube 1 was tested in a hydrogel phantom made of \SI{7}{\percent} polyvinyl alcohol and \SI{0.85}{\percent} phytagel solution, mimicking brain tissue properties. To advance the cannula, the actuation mechanism was connected to a linear rail with a 380:1 gear ratio brushed DC motor
(Pololu Robotics and Electronics (P/N 5227), NV, USA) attached to a precision lead screw that enables translation. As shown in Fig.\,\ref{fig:results}\,(h-1)-(h-3), the joint was able to steer through the hydrogel.

\section*{CONCLUSIONS AND DISCUSSION}
This work presents the fabrication of PI-based steerable joints using a femtosecond laser. The behavior of the joints was characterized, and a model to predict the joint angle for a given tendon stroke was validated. \textcolor{black}{Three different tube sizes were fabricated and characterized to show the material's behavior across various sizes.} The material showed a repeatable behavior after initial cycles. The steering of one of the designed joints in the hydrogel is shown. The method presented in this work can potentially be utilized to design PI-based joints for several applications in neurosurgery. \textcolor{black}{In comparison to nitinol-based joints, PI can potentially offer reduced interference in MRI and insulation for electrically-active end effectors.} In our future work, we aim to model the unloading behavior of the joint and integrate sensors for shape feedback to achieve precise navigation of the joint during neurosurgical interventions. 
\begin{table}[t]
\centering
\small
\caption{Joint Modeling Validation Error}
\label{table:rmse}
\vspace{-0.25cm}
\begin{tabular}{|c||c|c|c|}
\hline
& \textbf{Tube 1} & \textbf{Tube 2} & \textbf{Tube 3} \\
\hline
\hline
\textbf{RMSE Sample 1 ($^\circ$)} & 0.81 & 0.76 & 2.28\\
\hline
\textbf{RMSE Sample 2 ($^\circ$)} & 0.65 & 0.71 & 0.84\\
\hline
\end{tabular}
\vspace{-0.6cm}
\end{table}
\normalsize
\nocite{*}
\bibliographystyle{IEEEtran}
\bibliography{CRAS}

\end{document}